\documentclass[sigplan,screen]{acmart}

\usepackage{subcaption}
\usepackage{adjustbox}
\usepackage{multirow}
\usepackage{listings}
\usepackage[frozencache=true,cachedir=minted-cache]{minted}
\lstdefinestyle{rdfxml}{
    basicstyle=\ttfamily\scriptsize,
    commentstyle=\color{blue},
    morekeywords={rdf:RDF, rdf:Description, rdf:about, rdf:type, rdfs:label, rdfs:comment, kh-p:nextStep},
    sensitive=false,
    morestring=[b]",
    morecomment=[s]{<!--}{-->},
}

\AtBeginDocument{%
  \providecommand\BibTeX{{%
    \normalfont B\kern-0.5em{\scshape i\kern-0.25em b}\kern-0.8em\TeX}}}

\setcopyright{acmcopyright}
\copyrightyear{2018}
\acmYear{2018}
\acmDOI{XXXXXXX.XXXXXXX}

\acmConference[Conference acronym 'XX]{Make sure to enter the correct
  conference title from your rights confirmation emai}{June 03--05,
  2018}{Woodstock, NY}
\acmPrice{15.00}
\acmISBN{978-1-4503-XXXX-X/18/06}

\begin{document}


\title{Procedural Text Mining with Large Language Models}

\author{Anisa Rula}
\authornote{Both authors contributed equally to this research.}
\email{anisa.rula@unibs.it}
\affiliation{%
  \institution{University of Brescia}
  \streetaddress{Via Branze 38}
  \city{Brescia}
  \country{Italy}
  \postcode{25123}
}

\author{Jennifer D'Souza}
\email{jennifer.dsouza@tib.eu}
\affiliation{%
  \institution{TIB Leibniz Information Centre for Science and Technology}
  \streetaddress{Welfengarten 1B}
  \city{Hannover}
  \country{Germany}
  \postcode{30167}
}

\renewcommand{\shortauthors}{Rula and D'Souza}

\begin{abstract}
Recent advancements in the field of Natural Language Processing, particularly the development of large-scale language models that are pretrained on vast amounts of knowledge, are creating novel opportunities within the realm of Knowledge Engineering. In this paper, we investigate the usage of large language models (LLMs) in both zero-shot and in-context learning settings to tackle the problem of extracting procedures from unstructured PDF text in an incremental question-answering fashion. In particular, we leverage the current state-of-the-art GPT-4 (Generative Pre-trained Transformer 4) model, accompanied by two variations of in-context learning that involve an ontology with definitions of procedures and steps and a limited number of samples of few-shot learning. The findings highlight both the promise of this approach and the value of the in-context learning customisations. These modifications have the potential to significantly address the challenge of obtaining sufficient training data, a hurdle often encountered in deep learning-based Natural Language Processing techniques for procedure extraction. 

\end{abstract}




\maketitle

\section{Introduction}








Extracting complex knowledge from unstructured sources is a challenge: in the industrial domain, for example, troubleshooting documents may contain the description of long and articulated procedures (i.e., sequences of steps to be performed in a precise order and under specific conditions) and those natural language instructions may be represented in very different textual forms, thus making it hard for a knowledge extraction algorithm to correctly identify and structure the relevant information. Oftentimes, automatic extraction is followed by manual revision of domain experts. In any case, all machine-learning-based methods require training data which is often not readily available, therefore novel approaches are emerging to exploit interactive dialogues and language models~\cite{bellan2021process}.

Extracting procedural knowledge from human natural language instructions is a challenging task. 
Firstly, natural language instructions are not interpretable by machines. In the easiest case, the instructions are given as numbered lists which can easily be identified. However, complications arise when the document contains procedures in different forms: a list without numbers, an indented text or simply a full text in which the different steps are connected by conjunctions like "then", "afterwards", etc. Secondly, procedures can either be composed of only simple steps or contain other sub-procedures that are located elsewhere in the document. Thirdly, the procedures can differ substantially from one document to the other because of different authors' and editors' styles, but the goal would be to integrate information from different documents. 
Specifically, we investigate the potential of large language models (LLMs) in the context of extracting procedural knowledge from unstructured PDF documents. LLMs demonstrate remarkable capabilities in natural language processing, surpassing those possible using conventional symbolic AI and machine learning technologies \cite{mahowald2023dissociating}. Nevertheless, these models often lack knowledge of nuanced, domain-specific details and are susceptible to hallucinations.

This paper investigates the practical application of LLMs, with a particular focus on the advanced GPT-4 (Generative Pre-trained Transformer 4), to address the complex task of extracting procedures from unstructured PDF documents. The core of our research revolves around an incremental question-answering methodology, with a specific emphasis on harnessing LLMs in both zero-shot and in-context learning scenarios.
Our study is structured to encompass two distinct approaches to in-context learning. The first approach involves the incorporation of an ontology containing definitions and procedural steps, while the second approach integrates a limited dataset tailored for few-shot learning. This comprehensive investigation not only highlights the considerable potential of our chosen approach but also underscores the critical role played by customised in-context learning. These tailored modifications are poised to make significant progress in tackling a persistent challenge within deep learning-based NLP techniques: the scarcity of essential training data for procedure extraction.



Aligned with our exploration, three fundamental research questions guide our investigation:

\begin{itemize}
  \item \textbf{RQ1}: How well is ChatGPT4 able to list the steps and substeps in the text versus the ontology settings?
  \item \textbf{RQ2}: Is in-context learning beneficial for procedural text mining?  
  \item \textbf{RQ3}: Regardless of the ontology instantiation, is ChatGPT4 capable of the correct application of the ontology?
\end{itemize}

Through systematic empirical inquiry, this study not only contributes to the enhancement of procedural text mining but also offers insights into the capacity of in-context learning enhancements to surmount the constraints stemming from inadequate training data. The significance of this research extends beyond procedural extraction, resonating within the broader landscape of NLP applications and cultivating the evolution of more sophisticated and adaptable information retrieval systems.
Our code and dataset is publicly released \url{https://github.com/jd-coderepos/proc-tm/}.

The paper is structured as follows. We begin with a motivating example for our work in \autoref{sec:motiv-example}; then we describe the procedural ontology in \autoref{sec:ontology}. In \autoref{sec:approach}, we propose our approach and introduce the experimental dataset used for the task of procedural text mining in \autoref{sec:dataset}. Our experimental results are discussed both quantitatively and qualitatively next in \autoref{sec:results}. Finally a brief discussion on related work is offered in \autoref{sec:related} and concluding remarks in \autoref{sec:conclusion}.

\section{Motivating example}
\label{sec:motiv-example}

\begin{figure}[tb!]
\centering
\includegraphics[width=0.4\textwidth]{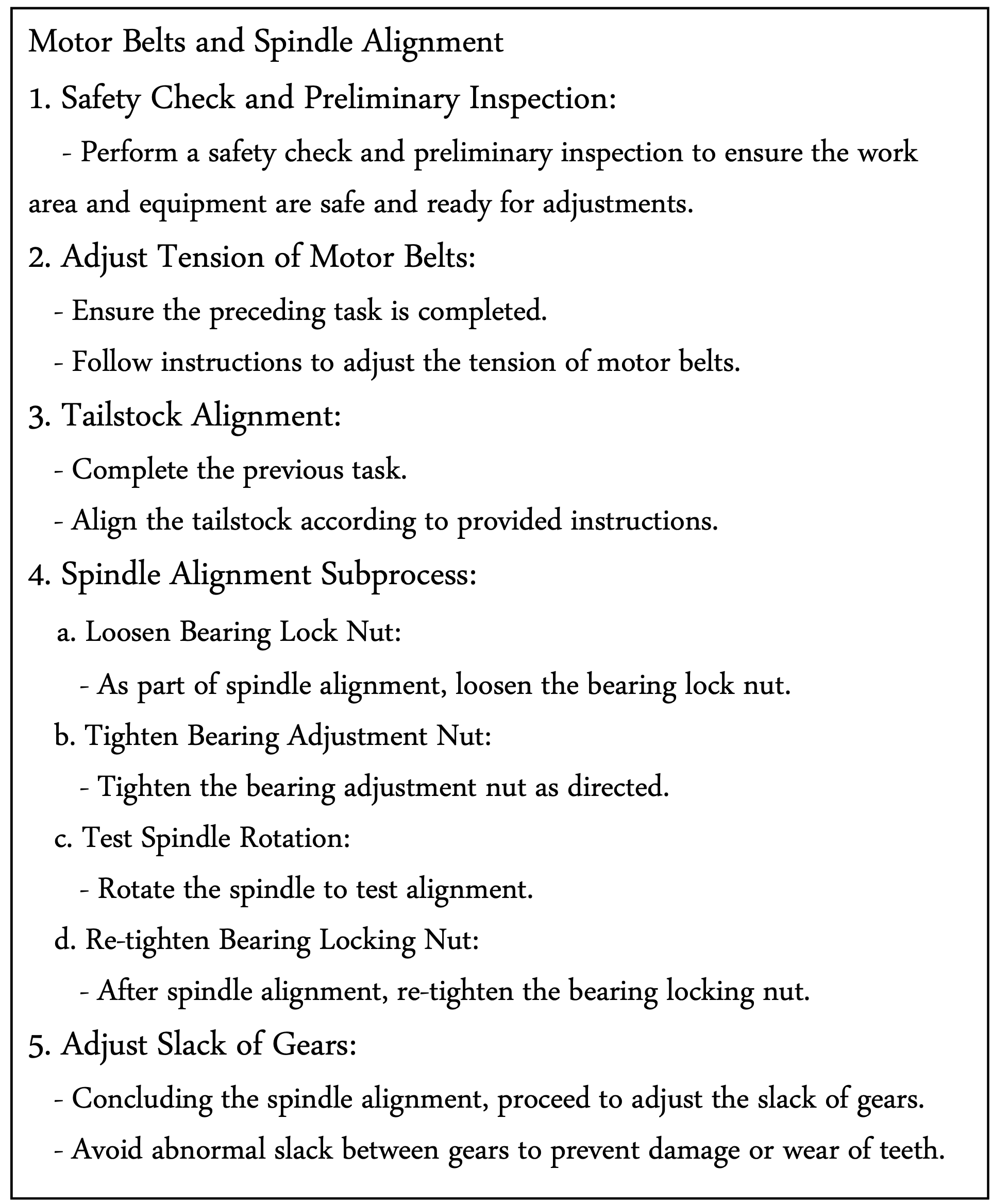}
\caption{Example of a procedure.}
\label{fig:procExample}
\end{figure}

Extracting the relevant procedure from the manuals becomes crucial in every context. Consider in a specific use case scenario a shop floor operator of the raw materials supplier who needs to follow a maintenance procedure with the gear head lathe machinery and thus he needs to interact with the maintenance manual. The procedure comprises a linear sequence of activities that must be completed in a specific order to address maintenance tasks and ensure optimal machine performance (see Figure \ref{fig:procExample}). 
The shop floor worker may need to answer some questions for which a simple keyword-based search in the document is not sufficient: \\
- \textit{What are the steps involved in performing routine maintenance on machinery gear head lathe?}\\
- \textit{Are there any sub-procedures or specialised steps for troubleshooting specific issues during maintenance procedures for machinery gear head lathe?}\\

To overcome the limitations posed by search methods reliant on keywords, it becomes necessary to extract and represent the procedural knowledge graph by means of a vocabulary that covers domain concepts and domain-specific terms used within procedures. This knowledge is extracted and represented with the support of an in-context learning strategy that enables the customisation of LLMs in a few-shot fashion. Their integration renders knowledge engineering accessible to individuals without specialised expertise in formal representation languages. This knowledge can be used to query semantic procedural data to answer queries that were not possible in the traditional unstructured form. Further, by querying enabled over semantically structured procedures, computers can now more intelligently assist others in efficiently managing, understanding, and executing procedures.


\section{Ontological representation of procedural knowledge} \label{sec:ontology}

In this section, we discuss our proposed vocabulary that translates aspects of procedures from unstructured text into equivalent machine-actionable structured representations. The Procedure vocabulary encompassed within the K-Hub Ontology\footnote{\url{https://knowledge.c-innovationhub.com/k-hub/}} plays a crucial role in defining the foundational concepts and relationships essential for the formal representation of procedures delineated within manuals \cite{RulaCABCBC23}. The vocabulary is designed to define the relationships and structure between plans and steps in a procedural context. It allows for specifying the sequence of steps within a plan, associating steps with plans, indicating the first and last steps of a plan, and decomposing steps into sub-plans. 
An explanation of the key elements is given below:

Properties
\begin{itemize}
    \item \texttt{p-plan:isStepOfPlan}, specifies that a step is part of a plan. It defines the relationship between steps and plans.
    \item \texttt{kh-p:nextStep}, indicates the sequence of steps within a plan. It associates a step with the next step that must be performed.
    \item \texttt{kh-p:startsWith}, associates a plan with its initial step. This indicates the first step in a plan.
    \item \texttt{kh-p:endsWith}, associates a plan with its last step. This indicates the final step in a plan.
    \item kh-p:isDecomposedAsPlan: Specifies that a step can be decomposed into a sub-plan. It associates a step with a plan that must be executed as a step of another plan.
\end{itemize}

Classes:
\begin{itemize}
\item \texttt{p-plan:Step}, represents the concept of a step in a procedure or plan. Each atomic activity is an instance of the \texttt{Step} concept. 
\item \texttt{p-plan:Plan}, represents the concept of a plan, which is composed of steps, which must be executed in a given order. A procedure is an instance of the \texttt{Plan} concept. 
\end{itemize}

\begin{figure}[tb!]
\centering
\includegraphics[width=0.4\textwidth]{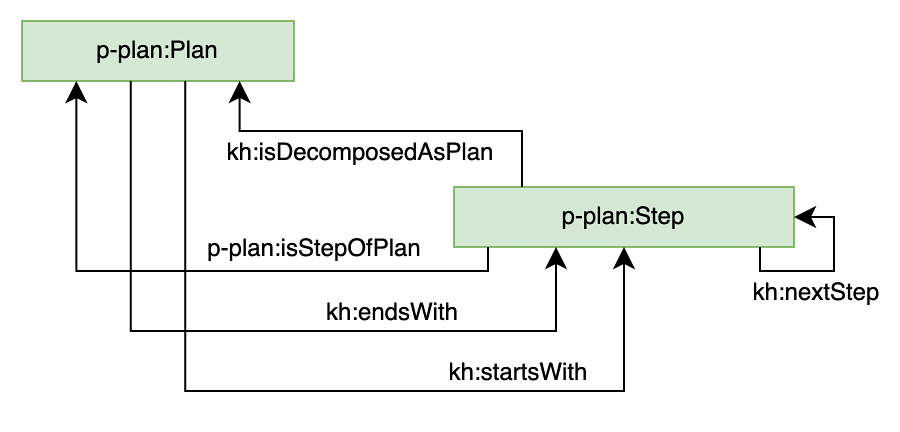}
\caption{Procedure ontology.}
\label{fig:procOnto}
\end{figure}

The ontology in Figure \ref{fig:procOnto} can be used to represent and query procedural knowledge in a structured and standardised manner, making it easier to manage and reason about procedural data. It aligns with the P-Plan ontology~\cite{garijo2012augmenting} and includes additional properties specific to procedural relationships.

Listing \ref{lst:procKG} RDF knowledge graph represents the "Motor Belts and Spindle Alignment" procedure as a plan composed of steps, including the relationships between steps and the plan's start and end points. It also defines a subprocess for spindle alignment within the main procedure.

\begin{lstlisting}[style=rdfxml, caption={RDF/XML Listing}, label={lst:procKG}]

<rdf:RDF xmlns:rdf="http://www.w3.org/1999/02/22-rdf-syntax-ns#"
         xmlns:rdfs="http://www.w3.org/2000/01/rdf-schema#"
         xmlns:kh-p="https://knowledge.c-innovationhub.com/k-hub/procedure#"
         xmlns:p-plan="http://purl.org/net/p-plan#">
    
    <!-- Step 11.3.3: Tailstock Alignment -->
    <rdf:Description rdf:about="kh-p:Step11_3_3">
        <rdf:type rdf:resource="p-plan:Step"/>
        <rdfs:label>Tailstock Alignment</rdfs:label>
        <rdfs:comment>Complete the previous task. Align the tailstock 
        according to provided instructions.</rdfs:comment>
        <kh-p:nextStep rdf:resource="kh-p:Step11_3_4"/>
        <kh-p:isStepOfPlan rdf:resource="kh-p-instance:Plan11_3"/>
    </rdf:Description>

    <!-- Step 11.3.4: Spindle Alignment -->
    <rdf:Description rdf:about="kh-p:Step11_3_4">
        <rdf:type rdf:resource="p-plan:Step"/>
        <rdfs:label>Spindle Alignment</rdfs:label>
        <rdfs:comment>A subprocess for spindle alignment.
        </rdfs:comment>
        <kh-p:isDecomposedAsPlan rdf:resource="kh-p:Plan11_3_4"/>
        <kh-p:isStepOfPlan rdf:resource="kh-p-instance:Plan11_3"/>
    </rdf:Description>

    <!-- Plan 11.3.4: Spindle Alignment Plan -->
    <rdf:Description rdf:about="kh-p:Plan11_3_4">
        <rdf:type rdf:resource="p-plan:Plan"/>
        <rdfs:label>Spindle Alignment Plan</rdfs:label>
        <kh-p:startsWith rdf:resource="kh-p:SubStep11_3_4_1"/>
        <kh-p:endsWith rdf:resource="kh-p:SubStep11_3_4_4"/>
    </rdf:Description>

    <!-- SubStep 11.3.4.1: Loosen Bearing Lock Nut -->
    <rdf:Description rdf:about="kh-p:SubStep11_3_4_1">
        <rdf:type rdf:resource="p-plan:Step"/>
        <rdfs:label>Loosen Bearing Lock Nut</rdfs:label>
        <rdfs:comment>As part of spindle alignment, 
        loosen the bearing lock nut.</rdfs:comment>
        <kh-p:nextStep rdf:resource="kh-p:SubStep11_3_4_2"/>
        <kh-p:isStepOfPlan rdf:resource="kh-p:Plan11_3_4"/>
    </rdf:Description>
</rdf:RDF>
\end{lstlisting}

\section{Approach} \label{sec:approach}






This section provides an overview of the methodology adopted to construct different versions of question-answering systems through in-context learning.
The primary focus during the design of conversational interactions was on the extraction of procedures, including their constituent steps and substeps as these form the fundamental components of any procedure. 
For the creation of the conversational system, we employed GPT-4 (Generative Pre-trained Transformer 4.0) hosted on ChatGPT-Plus, along with the \url{https://askyourpdf.com/} plugin for GPT-4. This choice was based on GPT-4  as a cutting-edge standard in LLMS. However, given that raw pre-trained language models may not seamlessly align with specific tasks, we employed in-context learning through prompting to customise the native model to varying extents. The following sections will provide detailed insights into the strategies adopted for formulating query templates and the customised models functioning as conversational systems.

\subsection{Prompting and questioning}
\label{prompting}

\begin{figure}[!h]
\includegraphics[width=0.5\textwidth]{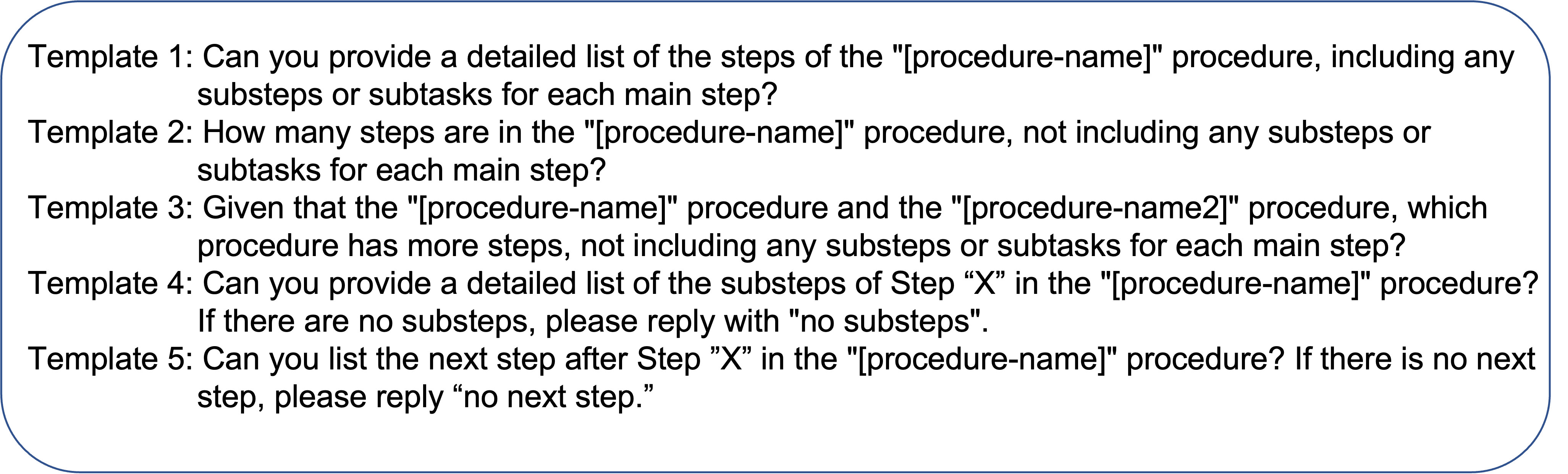}
\caption{Question templates and their ordering}
\label{fig:templates}
\end{figure}

Figure \ref{fig:templates} illustrates five distinct queries formulated for information extraction purposes. 
The questions, and therefore construction, are posed in an incremental manner. First, we ask questions about the list of steps and substeps and then we ask questions regarding aggregations and comparisons, and finally, the precedence relations among steps. We discuss shortly each of these templates:

\noindent \textit{Template 1 (List):} Requests a detailed list of steps for a specific procedure, including any step or substep for each main step. \\
\noindent \textit{Template 2 (Counting):} Asks for the total number of steps within a procedure. \\
\noindent \textit{Template 3 (Comparison):} Aims to identify the procedure with the maximum number of steps among a set of procedures. \\
\noindent \textit{Template 4 (Nested Procedures):} Asks about the presence of sub-procedures or substeps within the main procedure. \\
\noindent \textit{Template 5 (Sequence):} Asks for the next step after a specific step in a procedure and provides the step that comes immediately after the specified step.

These questions were structured incrementally to address the model's limitations in handling complex queries that encompass the entire procedural structure. This incremental approach shares similarities with the iterative process of crafting conceptual models often involving interactions with domain experts. It paves the way for versatile pipelines by combining diverse diverse incremental inquiries.

\subsection{In-context learning customisations}

"Contextual learning" involves training models to understand the contextual environment within which information is presented. This context can encompass surrounding text, images, or other data facets. Contextual learning is particularly important for tasks like NLP, where the meaning of a word or phrase can vary based on its surrounding context.

\textbf{Learning approaches: Raw vs Zero-shot vs 2-shot.} The initial learning approach aligns with "zero-shot learning." where the pre-trained GPT-4 model can generalise its knowledge to tasks it has not been explicitly trained on but gains its understanding by providing the model with some initial context. In contrast "2-shot learning" falls between traditional supervised learning and zero-shot learning. This approach involves training a machine learning model with just two labelled examples per class, enabling the model to generalise and make predictions for new instances even with limited labelled data.

\textbf{Contextual knowledge definitions.} Contextual knowledge is provided through the identification of the specific domain and intensional definitions of the procedure elements to be extracted. Relying on intensional definitions offers the advantage of compactness without requiring the provision of examples. These definitions consist of the main concepts and propteries defined in the ontology along with definitions of Procedure, Steps and Substeps. In the figure, each definition is labelled with the question it was used for. These choices were made to minimise external knowledge while providing an initial empirical assessment of using intentional definitions in the customisation of pre-trained models. 

\section{Our Dataset}
\label{sec:dataset}

To understand the effectiveness of our approach we select four domains
where public data are easily accessible and which cover all the challenges. 
\begin{itemize}
    \item Photography, manuals that provide instructions for operating, maintaining, and troubleshooting cameras, covering settings, capturing images or videos, lens care, and handling various scenarios.
    \item Manufacturing, manuals that provide instructions for production, quality control, assembly, maintenance, and safety in various manufacturing operations.
    \item Medicine, manuals that provide instructions for dental instrument usage, sterilisation, X-rays, oral hygiene instructions, and emergency protocols.
    \item Agriculture, manuals that provide instructions for the operation, maintenance, and safety protocols of farm equipment. 
\end{itemize}

We now give details on how we extracted the documents from each data source. First, we examine all the procedures defined in the manuals. Second, we specifically choose procedures that adhere to a structured format suitable for enumeration, such as numbering, bullet points, or clear indentation. Third, we prioritise procedures to be considered procedures that are only on one page or at most spanned across two consecutive pages. For each domain, we extract three examples of procedures either from the same manual or different manuals. 

\section{Results}
\label{sec:results}

In this section, we discuss the results obtained by leveraging ChatGPT4 for procedural text mining w.r.t. the 5 prompting scenarios introduced in \autoref{prompting}.

\subsection{Quantitative Evaluations}

The quantitative evaluations are applied only in the case of the first prompting scenario. In this case, ChatGPT4 is asked to extract all the steps including subtasks and substeps for a given procedure. The quantitative evaluations entail comparing all the extracted procedures versus their gold-standard annotations in text and ontologised formats. The results are reported in terms of ROUGE (Recall-Oriented Understudy for Gisting Evaluation) scores which are a set of evaluation metrics used to assess the quality of generated text by comparing them to the desired reference text.

For each of the four domains, roughly 3 or 4 procedures are considered, respectively, where the first prompting scenario is tested. These cumulative results are reported in \autoref{tab:results}. With this we answer \textbf{RQ1}, i.e. how well is ChatGPT4 ble to list the steps and substeps in the text versus ontology settings. As we see ChatGPT4 performs fairly high in the range above 70\% in extracting the procedures from the PDFs in the text format. The scores in the ontologized format are fairly low. In part, ChatGPT4 follows its own process for rewriting the extracted procedure in an ontology where it sometimes leaves out the use of prefixes and instead uses the full URL. Such differences in writing the output can cause the ROUGE scores to reflect low similarities between the extracted procedure and the human-written reference.

\begin{table}[!htb]
  \centering
  \begin{adjustbox}{width=1\columnwidth}
    \begin{tabular}{|p{1.25cm}|p{0.5cm}|p{1cm}p{1cm}p{1cm}p{1cm}|}
      \toprule
    \bf Domain & & \bf Rouge1 & \bf Rouge2 & \bf RougeL & \bf \stackbox[c]{Rouge-Lsum} \\
      \midrule
\multirow{2}{*}{\stackbox[c]{Photo-graphy}} & text & 79.6 &	74.4 &	79.4 &	79.6  \\ \cline{2-6}
     & ont. & 42.2 &	27.9 &	32.7 &	42.1 \\ \hline   
\multirow{2}{*}{Medicine} & text & 94.3 & 87.8 & 93.1 &	94.3  \\ \cline{2-6}
     & ont. & 47.5 & 38.8 & 43.1 & 47.5  \\ \hline 
\multirow{2}{*}{\stackbox[c]{Manufact-uring}} & text & 93.8 & 86.8 & 92.5 &	93.8  \\ \cline{2-6}
     & ont. & 55.7 & 46.0 & 52.9 & 55.5  \\ \hline  
\multirow{2}{*}{\stackbox[c]{Agri-culture}} & text & 75.4 & 66.5 & 74.3 &	74.4  \\ \cline{2-6}
     & ont. & 59.3 & 46.4 & 52.5 & 59.1  \\ \hline   
    \end{tabular}
     \end{adjustbox}
  \caption{Rouge scores comparing the automatically extracted procedures from the PDF manuals versus the gold-standard annotated procedures in text and ontologized (ont.) formats.}
  \label{tab:results}
\end{table}

\autoref{tab:results-2shot} reports the results for the 2-shot experimental setting for both the text and ontologised formats. With this, we answer \textbf{RQ2}, i.e. is in-context learning beneficial for procedural text mining. We see that it is. Our in-context learning 2-shot setting is as follows: ChatGPT4 is shown 2 gold-standard annotations as answers for the desired task where it is simply asked to comprehend the task, and the third prompt is where it is asked to perform the task. Thus the results reported in \autoref{tab:results-2shot} for the zero-shot setting is not identical to the results in \autoref{tab:results}. Nevertheless comparing the zero-shot/2-shot evaluations in the text format for the one example, we see that the results are boosted by at least 10 points in most cases. In the ontology setting, particular, the results are boosted by nearly 30 or more points. With this we see that ChatGPT4 acquires the skill of the desired application of the ontology when not just shown the ontology but also a few examples showing its desired application.

\begin{table}[!htb]
  \centering
  \begin{adjustbox}{width=1\columnwidth}
    \begin{tabular}{|p{1.25cm}|p{0.5cm}|p{1cm}p{1cm}p{1cm}p{1.1cm}|}
      \toprule
    \bf Domain & & \bf Rouge1 & \bf Rouge2 & \bf RougeL & \bf \stackbox[c]{Rouge-Lsum} \\
      \midrule
\multirow{2}{*}{\stackbox[c]{Photo-graphy}} & text & 85.5/85 &	77.6/77.6 & 85/85 & 85.5/85 \\ \cline{2-6}
     & ont. & 42.7/70.3 & 23.2/55.2 & 35.1/60.4 & 42.5/70.1 \\ \hline   
\multirow{2}{*}{Medicine} & text & 87.2/91.8 & 74.1/83.4 & 83.2/89.7 & 87.2/91.1 \\ \cline{2-6}
     & ont. & 44.8/86.9 & 36.3/83.5 & 40.4/85.8 & 44.6/86.9 \\ \hline 
\multirow{2}{*}{\stackbox[c]{Manufact-uring}} & text & 94.5/98.7 & 78.8/96.1 & 89.3/98.7 & 94.5/98.7 \\ \cline{2-6}
     & ont. & 59/78.9 & 51.5/69.4 & 56.6/73.8 & 58.3/78.1  \\ \hline  
\multirow{2}{*}{\stackbox[c]{Agri-culture}} & text & 74.7/92.9 & 62.5/79.2 & 73/92.2 & 74.1/92.2 \\ \cline{2-6}
     & ont. & 59.1/95.5 & 45.2/93.2 & 55.3/91.6 & 59.1/95.5 \\ \hline   
    \end{tabular}
     \end{adjustbox}
  \caption{Rouge scores comparing the automatically extracted procedures in the zero-shot/2-shot experimental settings from the PDF manuals versus the gold-standard annotated procedures in text and ontologized (ont.) formats.}
  \label{tab:results-2shot}
\end{table}

\subsection{Qualitative Evaluations}

The qualitative evaluations are discussed in terms of 9 observations presented as questions.

\textit{\textbf{1.} With the 2-shot in-context learning setting, can one actually tailor the model to respond in a certain way or a certain format?} E.g., for prompt scenario 5 in medicine, when asked for the next instruction in sequence, we would like the agent to reply with just the name or the sentence corresponding to the next instruction. However, the model in the "raw" setting seems to respond with the next instruction, but also with the substeps or additional information like notes or warnings related to the instruction. See \href{https://raw.githubusercontent.com/jd-coderepos/proc-tm/main/medicine/type5%20-%20sequence%20prompts/raw/chatgpt-response-example3.txt}{response} where it also adds the sentence ``Refer to page 17 of the document.'' which is additional information in the context of the instruction sentence ``CONTROL BOX INSTALLATION''. However, in the 2-shot setting, the model after seeing reference examples, responds with just the essential information for the same query as shown in-context. See \href{https://raw.githubusercontent.com/jd-coderepos/proc-tm/main/medicine/type5%20-%20sequence%20prompts/2shot/chatgpt-response.txt}{response} now reads ``... the next step after Step 4 "Head Installation" is Step 5 "Control Box Installation".''.

\textit{\textbf{2.} How well is the extraction of instructions handled when intermingled with images?} It seems that ChatGPT4 can distinguish between content related to text descriptions and images including their captions fairly well. For instance, procedure 1 in the agricultural domain for the prompt 1 i.e. to list all steps including substeps to the procedure about ``Operating the Hydrostatic Transmission'' presented such a case. To produce an accurate response, the agent had to overlook the image between step 1 and 2 which it successfully did. The \href{https://github.com/jd-coderepos/proc-tm/blob/main/agriculture/type1%20-%20list%20prompts/raw/chatgpt-response-example1.txt}{chatgpt response} can be compared to the \href{https://github.com/jd-coderepos/proc-tm/blob/main/agriculture/type1%20-%20list%20prompts/raw/goldstd-response-example1.txt}{gold-standard response}. A second successful example pertains to the ``Lowering ROPS Crossbar'' procedure in the following \href{https://github.com/jd-coderepos/proc-tm/blob/main/agriculture/manual-chunks/tractors-operating-example2.pdf}{manual} with two intermingled figures. Even though the \href{https://github.com/jd-coderepos/proc-tm/blob/main/agriculture/type1%20-%20list%20prompts/raw/chatgpt-response-example2.txt}{extracted response} from ChatGPT4 is somehow a paraphrased version of the \href{https://github.com/jd-coderepos/proc-tm/blob/main/agriculture/type1%20-%20list%20prompts/raw/goldstd-response-example2.txt}{gold-standard response}, it has successfully navigated the two interspersed images.

\textit{\textbf{3.} How does ChatGPT4 handle the extraction of a procedure across pages?} In the manufacturing domain, the instructions for "support plate installation" span two pages. This is example 2 handling the ``Support plate installation'' procedure in the set of \href{https://github.com/jd-coderepos/proc-tm/blob/main/manufacturing/type1%20-%20list%20prompts/raw/prompts.txt}{procedures}. The \href{https://github.com/jd-coderepos/proc-tm/blob/main/manufacturing/type1%20-%20list%20prompts/raw/chatgpt-response-example2.txt}{chatgpt response} for the type 1 prompt to list all steps, substeps etc. accurately reflected the expected \href{https://github.com/jd-coderepos/proc-tm/blob/main/manufacturing/type1%20-%20list%20prompts/raw/goldstd-response-example2.txt}{gold-standard}. Thus we see that the agent is able to assimilate information across pages while maintaining the right context. As another example from the manufacturing domain is example 4 in the context of type 1 prompt for "Removal and installation of Mechanical seal" sub-procedure for the "Shaft-seal maintenance" procedure. Here again, the \href{https://github.com/jd-coderepos/proc-tm/blob/main/manufacturing/type1%20-%20list%20prompts/raw/chatgpt-response-example4.txt}{chatgpt response} aside from splitting some instruction types, is 90\% in accordance to the \href{https://github.com/jd-coderepos/proc-tm/blob/main/manufacturing/type1%20-%20list%20prompts/raw/goldstd-response-example4.txt}{gold-standard}. It has successfully extracted the first step from the first page and the remaining steps from the next page of the manual. For the same instruction, when prompted in the 2-shot setting with examples of the desired response provided in the prompt, the \href{https://github.com/jd-coderepos/proc-tm/blob/main/manufacturing/type1%20-%20list%20prompts/2shot/chatgpt-response.txt}{chatgpt response} exactly matches the \href{https://github.com/jd-coderepos/proc-tm/blob/main/manufacturing/type1%20-%20list%20prompts/raw/goldstd-response-example4.txt}{gold-standard}. This behaviour is consistently observed for other domains as well. 

\begin{figure*}[!h]
\includegraphics[width=0.7\textwidth]{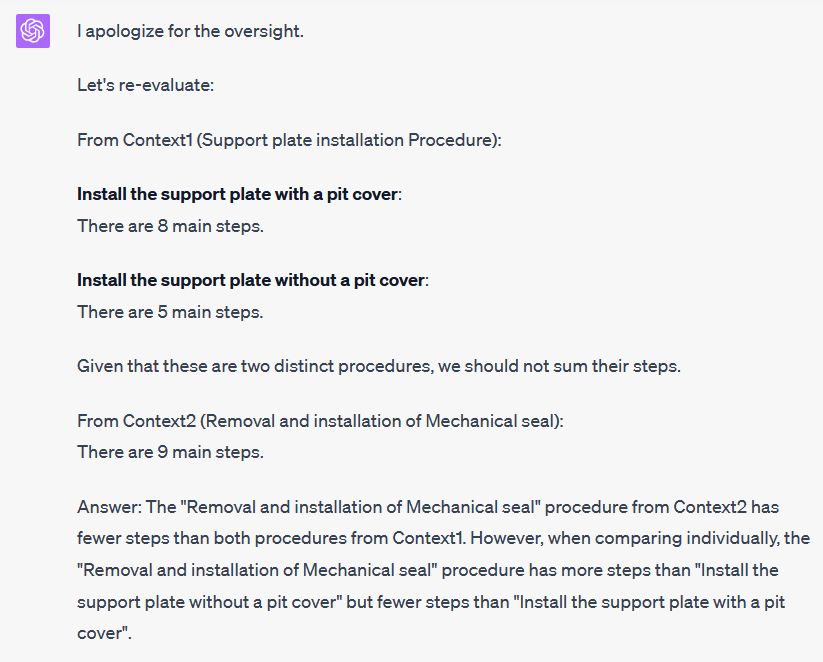}
\caption{Erroneous response from ChatGPT4 over a relatively simple mathematical reasoning task. The first sentence is correct: "Removal and installation of Mechanical seal" (9 main steps) $<$ both procedures from Context1 (8 main steps + 5 main steps). However the second sentence reflects the reasoning error: "Install the support plate with a pit cover" (8 main steps) > "Removal and installation of Mechanical seal" (9 main steps) > "Install the support plate without a pit cover" (5 main steps). Instead it should have been: the procedure in Context2 has more steps than either of the two procedures in Context1.}
\label{fig:gpt-err}
\end{figure*}

\textit{\textbf{4.} Apart from text generation discrepancies, has ChatGPT4 completely overcome the limitation of LLMs of not being truly capable of mathematical logic or reasoning but simply still relegated as very powerful statistical text generators~\cite{wei2022emergent,srivastava2022beyond}?} In the manufacturing domain, for type 3 comparison prompts, the language model was asked to compare the number of instructions given two contexts with procedures and reply which context had more instructions. Intriguingly, one of the contexts contained two procedures. Thus the task to the language model was to consider each independent procedure within each context and return which one had the most instructions. This is demonstrated in \autoref{fig:gpt-err}. The incorrect model response over a relatively simply reasoning task offers further credence to the conjecture: are large language models simply very good statistical generators and otherwise incapable of truly reasoning? 

\textit{\textbf{5.} Are our instructions completely unambiguous to the model?} We observed that in some cases they might be ambiguous. For instance, in prompt 4, i.e. probing the model to produce nested instructions setting, our prompt reads as follows:

\begin{table}[h!]
\centering
\begin{tabular}{ |p{8cm}| } \hline
Question: Can you provide a detailed list of the substeps of Step X in the given Context which refers to "[name]" procedure? If there are no substeps, please reply with "no substeps".
Answer:  \\ \hline
\end{tabular}
\vspace{-0.7cm}
\end{table}

Sometimes, if the main instruction itself is a rather long sentence, e.g., the generated step 7 \href{https://github.com/jd-coderepos/proc-tm/blob/main/manufacturing/type1%20-%20list%20prompts/raw/chatgpt-response-example4.txt}{here}, the prompt above proves ambiguous where ChatGPT4 splits the long instruction into a sequence of steps as in this \href{https://github.com/jd-coderepos/proc-tm/blob/main/manufacturing/type4%20-%20nested_proc%20prompts/raw/chatgpt-response-example3.txt}{response}. In-context learning alleviates ambiguity. In our 2-shot setting, the same prompt results with the \href{https://github.com/jd-coderepos/proc-tm/blob/main/manufacturing/type4%20-%20nested_proc%20prompts/2shot/chatgpt-response.txt}{correct response} 

\textit{\textbf{6.} Can ChatGPT4 effectively extract information from manuals in a 2-column format, processing each column accurately?}

We find that it can extract content to create a response cleanly column-by-column. However, if queried about a procedure described in the first column, it may not be able to detect the end of the procedure as relegated just to that one column. It could continue generating text even including a new procedure starting in the second column of the same page. E.g., the following \href{https://github.com/jd-coderepos/proc-tm/blob/main/agriculture/manual-chunks/tractors-operating-example1.pdf}{manual} on operating tractors has two distinct procedures, i.e. Operating the Hydrostatic Transmission (listed completely in column 1) and Using Cruise Control - 1026R (listed completely in column 2). In the prompt 1 scenario, ChatGPT4 when asked to list steps for ``Operating the Hydrostatic Transmission'' (see \href{https://github.com/jd-coderepos/proc-tm/blob/main/agriculture/type1%20-%20list%20prompts/raw/prompts.txt}{example 1}), it successfully extracts the relevant text but continues extracting text even for the ``Cruise Control'' procedure and lists its steps as substeps to the last step for ``Operating the Hydrostatic Transmission.'' See response \href{https://github.com/jd-coderepos/proc-tm/blob/main/agriculture/type1%20-%20list%20prompts/raw/chatgpt-response-example1.txt}{here}.

\textit{\textbf{7.} Is ChatGPT4 able to comprehend the correct application of the ontology even though the generated response does not match the gold-standard?} This is also our \textbf{RQ3}. 

A step in the ontology is described as follows: first given an instance name and initialised as a Step type of a Plan. E.g., ``kh-p-instance:Step2 a p-plan:Step ;'' Next the step is assigned a label. E.g., ``rdfs:label "Attach the hoses to the flowmeter ;''. Then the next step to the given step is specified. This can be a substep if the given step has substeps. E.g. ``kh-p:nextStep kh-p-instance:SubStep2\_1 ;''. Then the step in question is initialised as an instance of the corresponding main plan it belongs to. E.g., ``kh-p:isStepOfPlan kh-p-instance:Plan1 ;''. For those steps with substeps, the name of a subplan is specified. E.g., ``kh-p:isDecomposedAsPlan kh-p-instance:SubPlan2 .'' This subplan will be the plan to which substeps of the main plan are initialised. For a complete example, see lines 48 to 52 in the \href{https://github.com/jd-coderepos/proc-tm/blob/main/medicine/type1%20-%20list%20prompts/ontology/goldstd-response-example1.txt}{gold-standard example 1} in the medical domain. Note if a step does not have substeps then the specification of the next step goes to the next main step and not the substep. In addition, the line specifying the decomposed plan of a step will not be present. Now looking at the \href{https://github.com/jd-coderepos/proc-tm/blob/main/medicine/type1%20-%20list%20prompts/ontology/chatgpt-response-example1.txt}{ChatGPT4 response} for the same procedure, and the same instruction step in lines 17 to 27, we see that it has incorrectly specified the step in turn leading to an incorrect application of the ontology. First as a next step, instead of specifying the sub step, it used the subplan. Then for the substeps instead of specifying the decomposed plan for the main step at the step specification, it specifies it at the plan level which was initialised as the next step of the main step leading to something meaningless and not machine-actionable.

An incorrect application of the ontology is also found in \href{https://github.com/jd-coderepos/proc-tm/blob/main/medicine/type1%20-%20list%20prompts/ontology/chatgpt-response-example3.txt}{chatgpt response} to procedure 3 ``Installation of FM Type.'' Specifically take a look at the ``Pole Assembly Installation'' main step lines 12 to 23. The main step is initalised as a type of Plan. There is no specification of the decomposed plan with its substeps. Thus in the zero-shot setting ChatGPT4 cannot be expected to correctly apply the ontology. Promising enough, this changes in the 2-shot setting, where ChatGPT4 is showed two examples of the correct application of the ontology. Then via in-context learning, it is able to correctly apply the ontology. See the \href{https://github.com/jd-coderepos/proc-tm/blob/main/medicine/type1%20-%20list%20prompts/ontology%2B2shot/chatgpt-response.txt}{ChatGPT4 response} in the 2-shot setting for the same example, as a perfect application of the ontology, thereby showing that ChatGPT4 can be successfully guided via in-context learning toward the correct application of an ontology. Thus in a sense, at least for the ontology setting, it appears necessary to query ChatGPT4 via the in-context learning methodology showing it some examples with the correct application of the ontology.

As an observational note, in simpler ontology application scenarios, i.e. when there are just steps with no substeps, it does very well. 
E.g. from agriculture, for \href{https://github.com/jd-coderepos/proc-tm/blob/main/agriculture/type1%20-%20list%20prompts/ontology/prompts.txt}{example 4}, the \href{https://github.com/jd-coderepos/proc-tm/blob/main/agriculture/type1%20-%20list%20prompts/ontology/chatgpt-response-example4.txt}{ChatGPT4 ontologised response} for the prompt 1 scenario to list steps is almost  identical to the \href{https://github.com/jd-coderepos/proc-tm/blob/main/agriculture/type1%20-%20list%20prompts/ontology/goldstd-response-example4.txt}{gold-standard}.

\textit{\textbf{8.} Has ChatGPT4 hallucinated?} One needs to still be wary of the use of ChatGPT4 as it can still entirely hallucinate content. Consider the prompt 4 listing of nested procedures scenario, in the 2-shot setting, despite precise instructions as well as in-context example, when asked to list the substeps of step 3 for ``installation of the FM type'' procedure, ChatGPT4 still hallucinated the \href{https://github.com/jd-coderepos/proc-tm/tree/main/medicine/type4%20-%20nested_proc%20prompts/ontology%2B2shot}{entire response}. Compare this with the \href{https://github.com/jd-coderepos/proc-tm/blob/main/medicine/type1%20-%20list%20prompts/ontology%2B2shot/chatgpt-response.txt}{expected answer} for substep3\_1, 3\_2, and 3\_3. 

\textit{\textbf{9.} Is the 2-shot setting infallible, or does the model occasionally produce unexplained hallucinations?} In the 2-shot setting, we have also observed scenarios where the model has hallucinated text. While it may have grasped the ontology components and application relatively well, for reasons we found unexplainable the text generated as steps were entirely made up and could not be found in the manual. E.g., the generated \href{https://github.com/jd-coderepos/proc-tm/blob/main/manufacturing/type1%20-%20list%20prompts/ontology%2B2shot/chatgpt-response.txt}{ChatGPT4 response} in the 2-shot setting compared with the ontologised \href{https://github.com/jd-coderepos/proc-tm/blob/main/manufacturing/type1%20-%20list%20prompts/ontology/goldstd-response-example4.txt}{gold-standard} or text-based \href{https://github.com/jd-coderepos/proc-tm/blob/main/manufacturing/type1%20-%20list%20prompts/raw/goldstd-response-example4.txt}{gold-standard}.

\section{Related Work}\label{sec:related}
In prior research, it's vital to examine the methods used for procedural text mining and the incorporation of Large Language Models (LLMs) for knowledge extraction.

\textbf{Knowledge Extraction from Unstructured Sources}

Extracting complex knowledge from unstructured sources presents several challenges in several domains. This variability complicates the accurate extraction and structuring of relevant information through knowledge extraction algorithms which are usually applied to specific domains \cite{Jaradeh0SBA21}. The intricate nature of these documents requires manual review by domain experts after automated extraction, underscoring the limitations of machine-learning-based approaches \cite{DongPTMD21}. Innovative methods, including interactive dialogues and language models, have emerged to address the lack of readily available training data for machine learning methods \cite{RebmannA21,BertoliCFDGP22}. 

\textbf{Procedural Text Mining and Large Language Models (LLMs)}
In response to the challenges mentioned earlier, our research delves into the field of procedural text mining, capitalizing on advancements in Natural Language Processing (NLP). Large Language Models (LLMs) have emerged as a pivotal tool in this endeavor, surpassing the capabilities of traditional symbolic AI and machine learning technologies \cite{sun2023headtotail,bellan2022leveraging}. These models offer a means to address the intricate and extensive nature of procedural documents, with the potential to enhance knowledge extraction efficiency.

\textbf{Integration of LLMs in Knowledge Extraction}
Large Language Models (LLMs) demonstrate exceptional capabilities in natural language processing, surpassing what conventional symbolic AI and machine learning technologies can achieve \cite{mahowald2023dissociating}. These capabilities have sparked a substantial increase in proofs of concept and practical applications of LLMs, suggesting their potential utility in various knowledge-related tasks \cite{10.1145/3605943}. Nevertheless, the exploration of methods for effectively integrating LLMs into structured, controllable, and repeatable approaches for the development and deployment of such applications in production is still in its early stages and requires further detailed consideration \cite{pan2023unifying}.
Similarly, our study centers on the integration of LLMs, notably the state-of-the-art GPT-4 model, in the context of extracting procedural knowledge from unstructured PDF documents. 

\section{Conclusion}\label{sec:conclusion}

In this study, we explored the feasibility of employing in-context learning with pretrained language models to extract procedure elements from textual documents. We examined the native GPT-4 model and two customised variants, fine-tuned with procedure element definitions and limited examples.

Our findings suggest that in-context learning holds promise in addressing the challenge of training data scarcity in natural language processing for procedure element extraction. This approach opens up possibilities for initiating procedure element construction from natural language text, particularly in resource-constrained scenarios. Additionally, it underscores the importance of domain experts in refining model-generated information, emphasizing a human-in-the-loop approach.
This research has yielded valuable insights and lessons that will guide future investigations in this field.



\bibliographystyle{ACM-Reference-Format}
\bibliography{sample-base}

\end{document}